\documentclass{article}

\usepackage[final]{neurips_2022}

\usepackage[utf8]{inputenc} % allow utf-8 input
\usepackage[T1]{fontenc}    % use 8-bit T1 fonts
\usepackage{hyperref}       % hyperlinks
\usepackage{url}            % simple URL typesetting
\usepackage{booktabs}       % professional-quality tables
\usepackage{amsfonts}       % blackboard math symbols
\usepackage{nicefrac}       % compact symbols for 1/2, etc.
\usepackage{microtype}      % microtypography
\usepackage{xcolor}         % colors
\usepackage{graphicx}

\title{A Case for Business Process-Specific Foundation Models}

\author{%
  Yara Rizk \\
  IBM Research \\
  \texttt{yara.rizk@ibm.com} \\
  \And
  Praveen Venkateswaran \\
  IBM Research \\
  \texttt{praveen.venkateswaran@ibm.com} \\
  \AND
   Vatche Isahagian \\
   IBM Research \\
   \texttt{vatchei@ibm.com} \\
  \And
   Vinod Muthusamy \\
   IBM Research \\
   \texttt{vmuthus@us.ibm.com} \\
}

\begin{document}

\maketitle

\begin{abstract}
The inception of large language models has helped advance state-of-the-art performance on numerous natural language tasks. This has also opened the door for the development of foundation models for other domains and data modalities such as images, code, and music. 
In this paper, we argue that business process data representations have unique characteristics that warrant the development of a new class of foundation models to handle tasks like process mining, optimization, and decision making. These models should also tackle the unique challenges of applying AI to business processes which include data scarcity, multi-modal representations, domain specific terminology, and privacy concerns.
%In this paper, we argue that a business process is yet another information representation format that can leverage a foundation model to improve the performance of critical business process automation tasks like process mining, process optimization and decision making. 
 %, thereby preventing straightforward fine-tuning of existing models. We propose the need for a new class of foundation models tackling business process representations and these challenges, to improve the performance of important automation tasks like process mining, process optimization and decision making. 

%VI - Abstract changed
%In this paper, we posit that a similar renaissance is happening in the use of AI techniques in the business process domain and argue that business process data representations have unique characteristics that warrant the development of a new class of foundation models to handle tasks like process mining, optimization, and decision making. We also present a set of unique challenges associated with applying AI to business processes including data scarcity, multi-modal representations, domain specific terminology, and privacy concerns.
\end{abstract}

\section{Introduction}
Artificial intelligence, especially since the emergence of deep learning, has disrupted many areas of our lives from personal assistants like Alexa \citep{schneider2020users} to autonomous driving \citep{bernhart2016autonomous}. It has also been a disruptive force for businesses\footnote{https://www.gartner.com/smarterwithgartner/the-disruptive-power-of-artificial-intelligence}; 
deep learning is estimated to provide between \$3.5 trillion and \$5.8 trillion of annual value \cite{chui2018most} and can be the difference between companies' rise or demise. 
% Artificial intelligence (AI) has been projected to have over \$3.2 trillion in business value if adopted by businesses, according to Gartner \footnote{https://www.gartner.com/en/newsroom/press-releases/2018-04-25-gartner-says-global-artificial-intelligence-business-value-to-reach-1-point-2-trillion-in-2018}. 
%\footnote{https://www.mckinsey.com/capabilities/quantumblack/our-insights/most-of-ais-business-uses-will-be-in-two-areas}

In enterprise settings, business processes provide a structured framework for work. They define tasks, and identify their executors while capturing dependencies and providing logging and tracking capabilities. They also capture company policies and compliance with regulations. With many enterprises relying on the business process management paradigm to standardize their work, process management tools grew to a \$11.84 billion industry and is projected to grow to \$26 billion in 2028\footnote{https://www.marketwatch.com/press-release/business-process-management-market-size-growth-with-top-leading-players-growth-key-factors-global-trends-industry-share-and-forecast-2022-2031-2022-08-18}.

However, the existing landscape of work has been rapidly changing, requiring companies to move from their static business process practices to more agile and automated methods due to increased supply chain disruptions and skill shortages from the recent pandemic. 
%The changing landscape of work, including increased supply chain disruptions and skill shortages from the recent pandemic, has exacerbated the need to develop solutions to automate business process decision making. 
Thus, companies are making significant investments to adopt AI-driven tools for tasks like process prediction, visualization, translation, etc. \citep{mckendrick2021ai}, evidenced by the \$1+ million investments made by companies \footnote{https://www.gartner.com/en/newsroom/press-releases/2021-09-29-gartner-finds-33-percent-of-technology-providers-plan-to-invest-1-million-or-more-in-ai-within-two-years} 
and the projected \$3.2+ trillion business value produced by AI tech \cite{gartner2018}. 
Foundation models' recent success presents an opportunity to improve business process automation and management. 

Similar to natural language, images, or code snippets, business processes are yet another information representation paradigm. 
However, the unique and particular nature of process features and modalities can render existing foundation models inadequate to accurately understand and reason over them. 
Hence, developing successful foundation models for business process decision making requires research efforts to treat process data in a holistic manner instead of separate, independent modalities. 
% understand and determine approaches to represent process features, their modalities, relationships, and downstream tasks.

% % Possibly redundant
% The sudden outbreak of the COVID pandemic, has significantly changed the landscape of work, leading to the Great Resignation, skill shortages, supply chain disruptions, and working from home. As a result, companies are now increasing looking to consume AI techniques as part of their business process and automation tools. Most of these techniques involve training a Neural Network to perform a certain task (e.g. prediction, decision, etc.).
% %
% Like language or images or mathematics, processes are an information representation paradigm with particular properties that must be understood by neural networks if we hope to leverage foundation models to perform downstream tasks\footnote{Since processes also call individual nodes within a process a task, we will use ``downstream tasks'' to refer to foundation model specific prediction tasks and ``process tasks'' to refer to tasks within a process.}  
% like decision making within a process, process discovery, or process optimization.
% % End of possibly redundant

In this paper, we propose an approach to creating foundation models that factor in the complexity of process data. We also 
% In this paper, we provide an overview of business processes, their properties and the tasks that may be best suited for foundation models. We also propose an approach to training foundation models for the type of data encountered in business processes. Finally, we
discuss some of the challenges of creating foundation models for business processes and the risks and opportunities of foundation models' emergent behavior. First, however, we provide an overview of business processes, their unique properties and the tasks\footnote{Since processes also call individual nodes within a process a task, we will use ``downstream tasks'' to refer to foundation model specific prediction tasks and ``process tasks'' to refer to tasks within a process.} that may be best suited for foundation models.

\section{Background}

\subsection{Business Process Management}
\begin{figure}[tbp!]
    \centering
    \includegraphics[width=\linewidth]{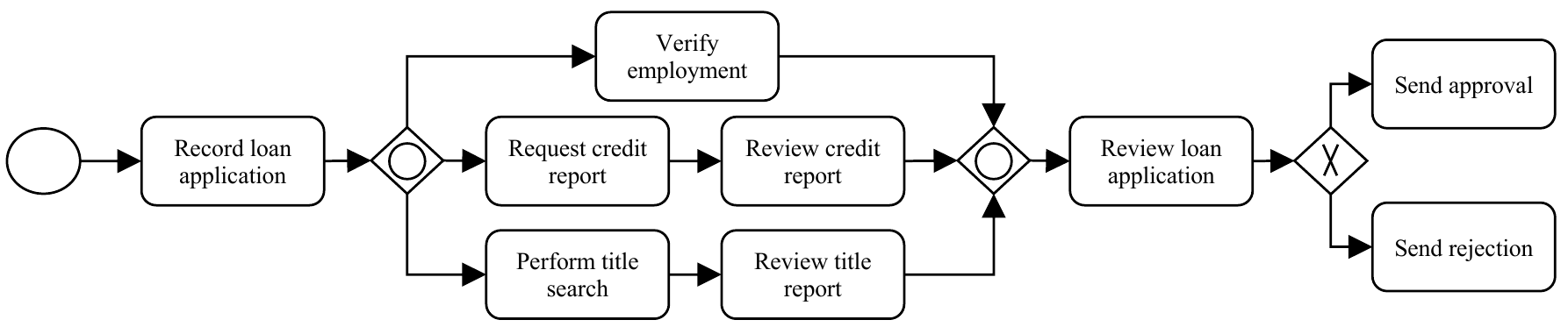}
    \caption{Example of a mortgage loan application process (Source: \cite{chakraborti2020robotic})}
    \label{fig:bp_loan}
\end{figure}

% - what is a business process
A \textit{business process} is a collection of ordered \textit{tasks}, followed by a business to produce a product or a service \cite{weske2012business}. Figure \ref{fig:bp_loan} shows the example of a mortgage application process where every application must go through the same steps before a decision is made. This allows mortgage lenders to structure their process, improve consistency across loan officers and track the execution of the process for accountability, auditing and improving the provided service. A graphical notation, known as business process model and notation (BPMN) \cite{grosskopf2009process}, is generally used to represent such processes, capturing the relationship between tasks (rectangular boxes with rounded edges) that must be completed by employee \textit{roles} within an organization, \textit{events} (circles) that can trigger processes or specific tasks within them, and \textit{decision points} (diamonds) that allow paths within the process flow to merge or diverge. %Connections between the various components determine the flow of information and actions (e.g., cycles are allowed). 
% Each task within the process can be performed by a specific employee \textit{role} within an organization. 
\textit{Swim lanes} are usually defined to place specific tasks within the scope of an employee role or department. A \textit{trace} is an execution of a process; each process can produce many distinct traces when executed depending on input events and other factors. 

% - business process management and automation and where AI fits in (applied/used)
Business process management consists of many problems related to the modeling or design, execution and governance of processes. 
Process mining or discovery analyzes event data to identify and derive processes from raw, unstructured data \cite{van2012process}. Ideally, process mining should produce a BPMN or similar representation for the discovered process. 
Process optimization or re-engineering looks to improve existing processes  \cite{arlbjorn2010business}. This requires making changes to the process representation while maintaining the properties that characterize a valid process. 
Conformance checking verifies that the ``as-is'' process (i.e., how the process is being executed in reality) does not deviate from the ``to-be'' process (i.e., how the process was theoretically designed to be executed) \cite{dunzer2019conformance}. 
Task automation through robotic process automation looks to create automation scripts that can programmatically execute tasks instead of humans \cite{van2018robotic}, whereas automation mining programmatically identifies the best tasks to automate  \cite{geyer2018process}. 
% Task automation mining combines process mining and task automation to identify which tasks within a process should be automated first to improve key performance indicators for companies \cite{geyer2018process}. 

% These are the topics I have in mind
% \begin{itemize}
%     \item Anomaly detection
%     \item prediction (next event,  next sub-sequence, remaining time)
%     \item Decisions
%     \item use of AI Planning to optimize process
%     \item Doc to process
% \end{itemize}

% -- tasks for BP don't have to just be decision making, we can be broader (like process mining, process optimization, automation, cogeneration, etc.) \\
% partial trace and predicting the rest of the trace --> can be considered decision making
% not just recommending approve or reject \\

\subsection{Foundation Models}
% - what are foundation models (how this came about from deep learning and big data) \\
Foundation models, coined in \cite{bommasani2021opportunities}, refer to deep neural network models trained on massive data and can be reused (with minimal modifications) for multiple downstream tasks. A key characteristic of foundation models is ``emergent'' knowledge: the model is able to make predictions and perform downstream tasks
that it has never seen before and wasn't trained on.  

% - examples of existing ones from other domains (NLP, computer vision, etc.) \\
Large language models were the first examples of foundation models; trained on billions of English sentences from the internet, the models learned the structure of language and became capable of performing natural language understanding and generation tasks \cite{devlin2019bert,brown2020language}. This has been followed by a wave of new foundation models catering to problems across different domains such as vision \citep{radford2021learning}, programming code, clinical and biomedical applications \citep{alsentzer2019publicly}, among others. 

% - how they are trained and used for prediction tasks \\
After training foundation models (generally) in an unsupervised or self-supervised paradigm, one of two approaches can be taken to use the model for a specific task. Either fine-tune on a small set of labeled data or create a prompt from labeled data to input to the model along with the input you want a prediction for. Both approaches have their pros and cons and have spurred many new open research questions and subfields of AI (e.g., prompt engineering \citep{liu2021pre}). 

\section{A Business Process Foundation Model}

\subsection{Overview}
The business process management literature is already rich with machine learning solutions to improve business processes (e.g. \cite{nguyen2022summarizing}). In recent years, there has been increased adoption of AI techniques (including natural language processing) in business processes for problems including predicting the next task or outcome of a process \cite{teinemaa2019outcome, evermann2016deep}, the remaining execution time of the process \cite{tax2017predictive, navarin2017lstm}, decision support \cite{agarwal2022process}, resource allocations \cite{zbikowski2021deep}, detecting drifts in process execution, and anomalous executions \cite{huo2021graph}. Declarative AI planning \cite{chakraborti2020d3ba} and reinforcement learning \cite{silvander2019business} have been used for process optimization. Natural language understanding has been used to declaratively extract process models \cite{aa2019extracting} and to provide conversational interfaces \cite{lopez2019process, rizk2020conversational}.

What all these problems have in common is a fundamental understanding of what a process is, its constituting components, its properties and its goals. 
However, the current narrow view of the literature when tackling these problems would lead to narrow solutions that may not realize the full potential of AI, especially when considering what foundation models could do. 
If we are able to encode this information in a foundation model, then we would be able to leverage this model to perform some of the tasks mentioned above. 

Foundation models for language learn the building blocks of language. There is a finite number of letters that words are made up of; not all letter sequences produce valid words. Sentences are composed of word sequences that must abide by the syntactic structures imposed by language. Words play specific syntactic and semantic roles within sentences and can have various semantic meanings based on context. Sentences also convey a semantic meaning that must be understood by the entity (person or otherwise) decoding the sentence. 
Similarly, foundation models for images learn that pixels with coordinates and values (in gray-scale or RGB or others) are combined to form lines that create shapes which have colors. An image has a foreground and a background; objects within an image have various spacial relationships with each other.

For business processes, foundation models need to learn about process artifacts, notation, and properties. %For instance, a process is composed of tasks and decision points; tasks have actors (in various roles) who perform those tasks within swim lanes; one execution of a process produces a trace and each process is associated with many traces. 
Furthermore, intra- and inter-process features have been shown to have an effect on the prediction \cite{senderovich2017intra}. Once a deep learning network internalizes all these concepts, then we can start performing more complex downstream tasks that rely on this foundational understanding like optimizing processes or discovering them from unstructured data and events.%, as shown in Figure \ref{fig:foundation_plus_verticals}.

% \begin{figure}[tbp!]
%     \centering
%     \includegraphics[width=0.7\linewidth]{stack}
%     \caption{Foundational knowledge versus downstream tasks in business process}
%     \label{fig:foundation_plus_verticals}
% \end{figure}

% - what tasks can benefit from foundation models \\
% - are there any existing foundation models we can use (this may be better for section 3) \\

% - discuss what types of data and their properties exist in BPA (look at the RPA Forum 2020 draft) \\
% - discuss what downstream tasks would fall within the scope of the foundation model \\
% - discuss possible design decision considerations for the foundation model (e.g., type of network, training hyperparameters, etc.) \\
% - discuss the challenges of training a foundation model for BPA that may be different from other domains (e.g., multi-modal data, not a lot of data? privacy concerns, breadths of the tasks?) \\
% -- possible solution: federated learning and secure/data preserving training \\
% - data augmentation, data generation strategies \\
% -- computational time for predictions at run time \\

% challenges:
% How can we train the foundation model to incorporate business acronyms.
% What incentives do business have to train (even if there is a privacy preserving Federated Learning approach).

% vatche and vinod's paper on few shot learning can be added here

\subsection{Data Types in Business Processes}
The data describing business processes and generated from their execution consists of many different types of data. Whether we consider business process data to be a new modality in machine learning \cite{chakraborti2020robotic} or treat it as a multi-modal problem, we first need to understand what types of data exist before we can effectively train foundation models. 

The first type of data embodied in a business process is a graph which represents the control flow of a process where tasks and decision points are connected to form a directed graph with cycles, branches, root nodes and end nodes \cite{white2004introduction}.
Once a process is executed, a sequence of events is generated, referred to as a process trace. One process may have many different traces representing the various traversals of the graph and different decisions at decision points. 

Processes also have metadata associated with the process and with the events within a process which are generally represented by a multi-dimensional set of attributes that can be binary, categorical or continuous. For example, each task in the process is typically associated with a human worker (e.g., loan officer, claims processor) from the enterprise organization, who are geographically distributed, have different working timezone, vacation, and holiday schedules. These human workers cannot work on two process cases at once, which in turn creates an implicit limit on the number of associated concurrent tasks across process instances. Events and tasks within a process can also have unstructured documents associated with them (e.g., images, text, video, audio). Interactions between participants (including social networks) in a process can be represented by graphs and times series data. 

Considering only a subset of data would provide an incomplete view of the business process and may lead to sub-optimal predictions by machine learning models. Thus, it is important to identify effective approaches to handle these types of complex applications and interactions with diverse data types. 

%\todo{Say something about additional semantics, such as people being assigned to tasks and metadata about people (roles, skills, org chart, vacation schedules, time zones); notion of cases or process instances; and derived inter-case features, such as people not being able to work on two cases at once, which has implicit limits on the number of associated concurrent tasks across process instances. We can make the argument that a language model might not be the right fit to encode these aspects.}

%\todo{Should we also mention things like policy documents that define guardrails for the process? These are sometimes structured, but often unstructured.}

\subsection{Downstream Tasks}
We distinguish between two types of downstream tasks for foundation models: domain agnostic vs. domain specific. Domain agnostic downstream tasks can be process mining, process optimization, trace prediction, etc. Domain specific downstream tasks can be process task prediction, decision recommendation at a decision point in a process, automation of process tasks, etc. 
Furthermore, some of these downstream tasks can be time sensitive vs. not. For example, identifying a process from unstructured data can be performed offline. However, a decision making step during process execution is more time sensitive; the foundation model needs to make a decision within seconds or minutes (possibly) as opposed to hours or days. 
Depending on the type of downstream tasks, we may need different versions of foundation models (e.g., computational heavy vs. light-weight). 

\subsection{Model Architecture}

\begin{figure}[tbp!]
    \centering
    \includegraphics[width=0.9\linewidth]{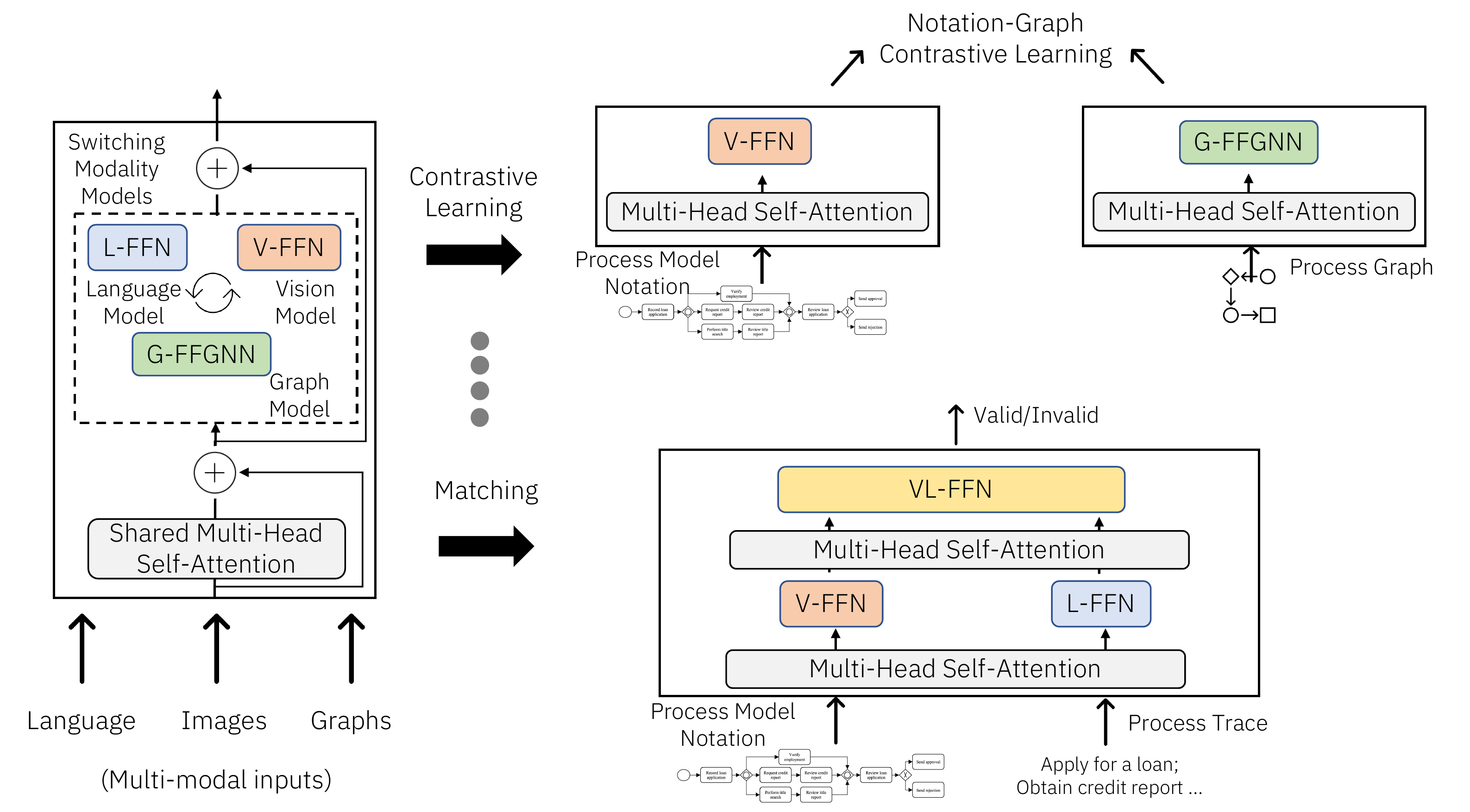}
    \caption{Using Mixture of Modality Experts (MoME) Transformer to pre-train a Business Process Foundation Model on different tasks}
    \label{fig:transformer_architecture}
\end{figure}

Past work on pre-training foundation models in multi-modal settings have focused on Vision-Language tasks. They learn cross-modal representations to align information using approaches like contrastive learning, matching, masked modeling, etc \citep{radford2021learning,kim2021vilt,yu2022coca}. 
These efforts primarily use one of two architectures. 
The first is a \textit{dual-encoder} architecture to encode different data modalities separately, and then use cosine-similarity of the feature vectors for modality interaction. This shallow interaction between the different modalities has been shown to perform poorly on several tasks \citep{jia2021scaling}. 
The second is a \textit{fusion-encoder} architecture with cross-modal attention, to jointly encode all possible data pairs to compute similarity scores for tasks. This results in a quadratic (for two modalities) time complexity and much slower inference speed than dual-encoder models whose time complexity is linear. 

We envision leveraging a recent approach, called Mixture-of-Modality-Experts (MoME) \citep{wang2021vlmo,wang2022image}, that uses a pool of modality models to replace the feed-forward network in a standard transformer architecture. It switches between different modality models to capture modality-specific information, and then uses shared self-attention across modalities to align information. Figure \ref{fig:transformer_architecture} describes our vision for implementing MoME for business process tasks, where we define expert feed-forward network (FFN) and feed-forward graph neural network (FFGNN) models for different modalities (language, vision, graph) and their combinations. 

Depending on the modality of the input vectors, the transformer selects the appropriate mixture of expert models to process the input. For instance, if the input consists of vectors representing process traces and process model notations, the transformer would pick the language and vision models to encode the inputs and a vision-language model to capture more modality interactions. Traditional pre-training tasks like contrastive learning, masking, matching, etc, can be performed to capture cross-modal information in the business process context, and we show an example of two pre-training tasks in Figure \ref{fig:transformer_architecture}.

\section{Challenges}

\subsection{Data Scarcity and Privacy Concerns}
A majority of foundation model training efforts consider tasks involving the generation of natural language (e.g., OpenAI GPT-3, Google T5), images (e.g., the recent DALL·E 2 model) or code syntax (e.g., GitHub Copilot, Amazon CodeWhisperer). An inherent advantage of these tasks, is the prevalence of a variety of relevant and labeled/unlabeled training data that have been collected and open-sourced by the larger research community. 

However, for business processes, there is a lack of sufficient labeled open-source real-world data to train foundation models. A big reason for this, is the inherent proprietary nature of business processes, resulting in most corporations being unwilling to share their data and models. While there have been some efforts towards democratizing business process data, such as the Business Process Intelligence (BPI) Challenges\footnote{https://www.tf-pm.org/newsletter/newsletter-stream-2-05-2020/bpi-challenges-10-years-of-real-life-datasets}, they have also stated the growing difficulties in obtaining real-world data from corporations, citing privacy concerns. 

Therefore, enabling foundation models for business processes would require addressing the critical challenge of data availability. This could entail solutions involving privacy-preserving training such as federated learning approaches, wherein models can be trained on data across multiple business units and corporations without involving any data sharing. Other possible solutions could involve data generation and augmentation techniques to leverage patterns from the literature (e.g., insertion of new tasks in the process, optionalization of tasks that were previously required in the process, and resequentialization of tasks \cite{maaradji2017detecting}) or ones existing in the data to create realistic new process data . Generative models (e.g., GANs) could be used to create new data instances hallucinated from existing processes. However, such approaches would also require crowd-sourced data validation (by subject matter experts) and labeling efforts to ensure training data quality. 

%\todo{Also mention data augmentation techniques, such as dropping activities in a process, reordering them, etc. There was a process drift paper that had a taxonomy of modifications. (Vatche this is one of the papers that Austin used for his internship work.) --DONE}

\subsection{Breadth of Tasks}

As with other domains, business process mining, monitoring, and automation can comprise of a multitude of possible tasks. These could be (1) process predictions -- such as predicting a future process sequence given a partial trace, process completion time, process failures, etc., (2) process synthesis -- including synthesising new process models from specifications or natural language input, process visualizations, etc., (3) explainability and summarization -- wherein models are expected to explain various business process decisions and predictions, as well as provide accurate summaries of process traces, among other tasks. 

These tasks involve different data modalities and input/output structures. Some of these tasks operate using text, some with images, and others with graphs. Hence, training a singular foundation model across these different tasks is a significant challenge. Determining the appropriate model parameters in this situation would require techniques like meta-learning \citep{finn2017model} to ensure minimal additional training to perform well on different downstream tasks. 

There have been several recent research efforts to incorporate multi-modality in foundation models. For example, \cite{zeng2022socratic} propose a prompt-driven approach to combine language, vision and audio models in a symbiotic manner to exchange information with each other and capture multi-modal knowledge. Similarly, \cite{wang2022image} propose a multi-modal foundation model \texttt{BEIT-3} for vision and language tasks, that uses a multiway transformers network to align various modalities. However, these models do not work for many business process automation tasks, thereby requiring a new foundation model initiative. 

\subsection{Domain Specific Language}

Tasks based on natural language have well-defined language constructs and semantic meaning for models to reason on. However, business processes often have acronyms and technical phrases which are not common knowledge, but are critical for the model to understand. Additionally, process models often adhere to different standards and graphical notations such as the Business Process Model and Notation (BPMN), Decision Model and Notation (DMN), Case Management Model and Notation (CMMN), etc.  \citep{white2004introduction,wiemuth2017application}. 

Hence, it is a critical challenge to develop a domain specific language (DSL) to enable foundation models to reason over such business process specific terminology. In addition, such a DSL would also enable users to enforce business policies and ensure the validity of the model outputs using techniques such as constrained semantic decoding \citep{poesia2022synchromesh}. However, the number of business domains and terminology is ever-increasing and nearly impossible to fully capture. This would result in situations where the model has limited knowledge or information, reflecting zero-shot or few-shot settings, that would require approaches like prompt-based fine-tuning of the model. 

\subsection{Prompt Engineering for Business Processes}

Many real-world tasks may have very little, or no data available to fine-tune foundation models. However, the use of prompts and in-context examples have been shown to enable language models to perform significantly well in zero-shot and few-shot settings \citep{radford2019language, brown2020language, sanh2021multitask}. The popularity of language tasks has even resulted in a public repository \citep{bach2022promptsource} of natural language prompts. 

While the use of prompts has demonstrably improved performance, foundation models have also been shown to be extremely sensitive to prompt engineering. For instance, \cite{zhao2021calibrate} and \cite{min2022rethinking} have shown that small changes to the prompt such as changing the prompt structure, reordering, and even the number of examples, can result in a significant drop in model performance. They also demonstrate how model biases arising from the pre-training data, can impact its performance when fine-tuned for downstream tasks. 

This presents several challenges for business process models. Firstly, while the structure of prompts may often be straightforward for language tasks (e.g., questions for question-answering), this is not the case for many business process tasks. For instance, tasks involving the translation of natural language specifications to process models or summarizing process models using text, would require careful prompt engineering. Prompts in the business process domain can involve images or even graph structures, and identifying the most relevant examples or prompts also presents a challenge.  
Moreover, ensuring the robustness of the model to biases during the pre-training process is critical. 

\subsection{Human-in-the-loop Feedback and Model Robustness}

Many process automation tasks involve critical decision-making steps. 
The sensitive and regulated nature of business domains often results in the requirement of human feedback to be present as part of the decision making pipeline. This feedback could involve the enforcement of corporate policies, ensuring the validity of model outputs, changes to intermediate decisions of the process pipeline, among others. 
Hence, process models would require an optimized approach to incorporate such human-in-the-loop feedback. Since fine-tuning large foundation models is an expensive process, it may not always be possible to continually update the model parameters with user feedback, thereby requiring approaches to incorporate the feedback within subsequent input prompts. 

Additionally, the influence of malicious actors and data biases on model decisions can have a significant and costly impact on businesses. For instance, adversarial prompts and feedback could be used to bias the model to output incorrect or inappropriate decisions, or even obtain any confidential information used to train or fine-tune the model \citep{bommasani2021opportunities, carlini2021extracting}. Hence, approaches to improve model robustness are critical for business process tasks. For instance, coupling constrained decoding with model outputs, where businesses can explicitly specify guardrails or policies \citep{rizk2022can}, and careful consideration of data biases, distribution shifts, and information leakage during the pre-training process are important. 

\section{Risks, Opportunities, and Next Steps}
The emergent behavior of foundation models has been a point of intrigue and concern in various fields like healthcare \cite{wiggins2022opportunities} and education \cite{blodgett2021risks}. For, business processes, things are no different. On the one hand, as foundation models become capable of generating, modifying and executing parts of a process, concerns around violating industry standards or company policies, auditability and interpretability must be addressed to ensure wide-spread adoption.  
% One of the immediate considerations is the possibility of violating company policies or industry regulations; as foundation models are used to make decisions within a process or design and optimize processes, they should have the ability to factor in regulations, standards or policies that may be encoded as rules within industries or companies. Furthermore, auditability is critical in many of the industries that rely on business processes; factoring in this requirement when developing a framework to integrate foundation models into the business world becomes indispensable. Further, insuring a human-in-the-loop (who is not a machine learning expert) can interact and understand the interface with foundation models is just as critical. 
%
% While these are only a few of the risks of adopting foundation models in business process management, we also believe that there are many benefits. For example,
On the other hand, using generative models to produce new business processes can unlock tremendous optimizations and new ways to do work that can help business achieve profitability without sacrificing sustainability and environmental impact. Also, foundation models can help make data driven decision making a reality for business processes. 

In summary, we believe that foundation models for business processes have tremendous potential to advance the field of process management and integrate AI into their practices. Both AI and BPM communities need to join forces to create the proper infrastructure to train and use such foundation models. Next steps for the community include identifying existing data sources and curating specialized datasets for training and fine-tuning. Safeguards should also be put in place to ensure that foundation models' emergent behavior does not have negative side-effects that may hinder its adoption in industry.

% Bibliography
\bibliographystyle{plainnat}
\small
\bibliography{refs}

\end{document}